\documentclass{article}

\usepackage{arxiv}

\usepackage[utf8]{inputenc} 
\usepackage[T1]{fontenc}    
\usepackage{hyperref}       
\usepackage{url}            
\usepackage{booktabs}       
\usepackage{amsfonts}       
\usepackage{nicefrac}       
\usepackage{microtype}      
\usepackage{lipsum}		
\usepackage{graphicx}
\usepackage{doi}

\usepackage{caption}
\usepackage{subcaption}
\usepackage{amsmath}

\usepackage{graphicx}
\usepackage{booktabs}

\usepackage{multirow}
\usepackage[normalem]{ulem}
\usepackage{arydshln} 
\usepackage{xcolor}
\usepackage{listings}
\usepackage{xcolor}
\newcommand{\review}[1]{\textcolor{black}{#1}}

\definecolor{codegreen}{rgb}{0,0.6,0}
\definecolor{codegray}{rgb}{0.5,0.5,0.5}
\definecolor{codepurple}{rgb}{0.58,0,0.82}
\definecolor{backcolour}{rgb}{0.95,0.95,0.92}

\lstdefinestyle{mystyle}{
    backgroundcolor=\color{backcolour},   
    commentstyle=\color{codegreen},
    keywordstyle=\color{magenta},
    numberstyle=\tiny\color{codegray},
    stringstyle=\color{codepurple},
    basicstyle=\ttfamily\footnotesize,
    breakatwhitespace=false,         
    breaklines=true,                 
    captionpos=b,                    
    keepspaces=true,                 
    numbers=left,                    
    numbersep=5pt,                  
    showspaces=false,                
    showstringspaces=false,
    showtabs=false,                  
    tabsize=2
}

\title{Spatial Context-based Self-Supervised Learning for Handwritten Text Recognition}

\author{ {Carlos Penarrubia}\\
	University of Alicante\\
	\texttt{carlos.penarrubia@ua.es} \\
	\And
        {Carlos Garrido-Munoz} \\
	University of Alicante\\
	\texttt{carlos.garrido@ua.es} \\
	\And
        {Jorge Calvo-Zaragoza} \\
	University of Alicante\\
	\texttt{jcalvo@dlsi.ua.es} \\
        \And
        {Jose J. Valero-Mas} \\
	University of Alicante\\
	\texttt{jjvalero@dlsi.ua.es} \\
}

\renewcommand{\shorttitle}{\textit{arXiv} Template}

\hypersetup{
pdftitle={A template for the arxiv style},
pdfsubject={q-bio.NC, q-bio.QM},
pdfauthor={David S.~Hippocampus, Elias D.~Striatum},
pdfkeywords={First keyword, Second keyword, More},
}

\begin{document}
\maketitle

\begin{abstract}
Handwritten Text Recognition (HTR) is a relevant problem in computer vision, and implies unique challenges owing to its inherent variability and the rich contextualization required for its interpretation. Despite the success of Self-Supervised Learning (SSL) in computer vision, its application to HTR has been rather scattered, leaving key SSL methodologies unexplored. This work focuses on one of them, namely Spatial Context-based SSL. We investigate how this family of approaches can be adapted and optimized for HTR and propose new workflows that leverage the unique features of handwritten text. Our experiments demonstrate that the methods considered lead to advancements in the state-of-the-art of SSL for HTR in a number of benchmark cases.
\end{abstract}

\keywords{Self-Supervised Learning \and Handwritten Text Recognition \and Spatial Context}

\section{Introduction}
\label{sec1}
Handwritten Text Recognition (HTR) is the research area in the field of computer vision whose objective is to transcribe the textual content of a written manuscript into a digital machine-readable format~\cite{VidaToselliRiosVilaCalvoZaragoza:PR:2023}. This area not only plays a key role in the current digital era of handwriting by electronic means (such as tablets)~\cite{BezerraZanchettinToseeliPirlo:Book:2017}, but is also of paramount relevance for the preservation, indexing and dissemination of historical manuscripts that exist solely in a physical format~\cite{muehlberger2019transforming}.

HTR has developed considerably over the last decade owing to the emergence of Deep Learning \cite{nikitha2020handwritten}, which has greatly increased its performance. However, to attain competitive results, these solutions usually require large volumes of manually-labelled data, which constitutes their main bottleneck. One means by which to alleviate this problem, Self-Supervised Learning (SSL), has recently gained considerable attention from the research community~\cite{ozbulak2023know}. SSL employs a so-called \textit{pretext} task to leverage collections of unlabelled data for the training of neural models to obtain descriptive and discriminative representations~\cite{balestriero2023cookbook}, thus reducing the need for large amounts of labelled data.

The pretext tasks can be framed in different categories according to their working principle~\cite{jing2020self, ozbulak2023know}, with the following being some of the main existing families: (i) {\it image generation} strategies~\cite{liu2021self}, which focus on recovering the original distribution of the data from defined distortions or corruptions; (ii) {\it discriminative learning} methods~\cite{jaiswal2020survey}, whose objective is to learn representative and discernible codifications of the data. In the last one, we can find {\it spatial context} methods, which focus on either estimating geometric transformations performed on the data~\cite{gidaris2018unsupervised}---\emph{i.e.} spatial information---or inferring the disposition of a set of patches in which the image is divided as a puzzle~\cite{doersch2015unsupervised,noroozi2016unsupervised}---\emph{i.e.} structural information.

Nevertheless, the development of SSL in computer vision has predominantly focused on image classification, whereas other domains such as HTR have garnered noticeably less attention. In this respect, although there are several research efforts that explore ad-hoc adaptations from strategies used in classification tasks~\cite{aberdam2021sequence,zhang2022chaco} or new methods particularly devised for these cases \cite{luo2022siman,souibgui2023text}, there is a remarkable scarcity of proposals, thus \review{leaving substantial SSL areas unexplored hindering the development of the HTR field.}

\begin{figure}[t]
  \centering
  \begin{subfigure}{\linewidth}
    \centering
    \includegraphics[width=0.5\linewidth]{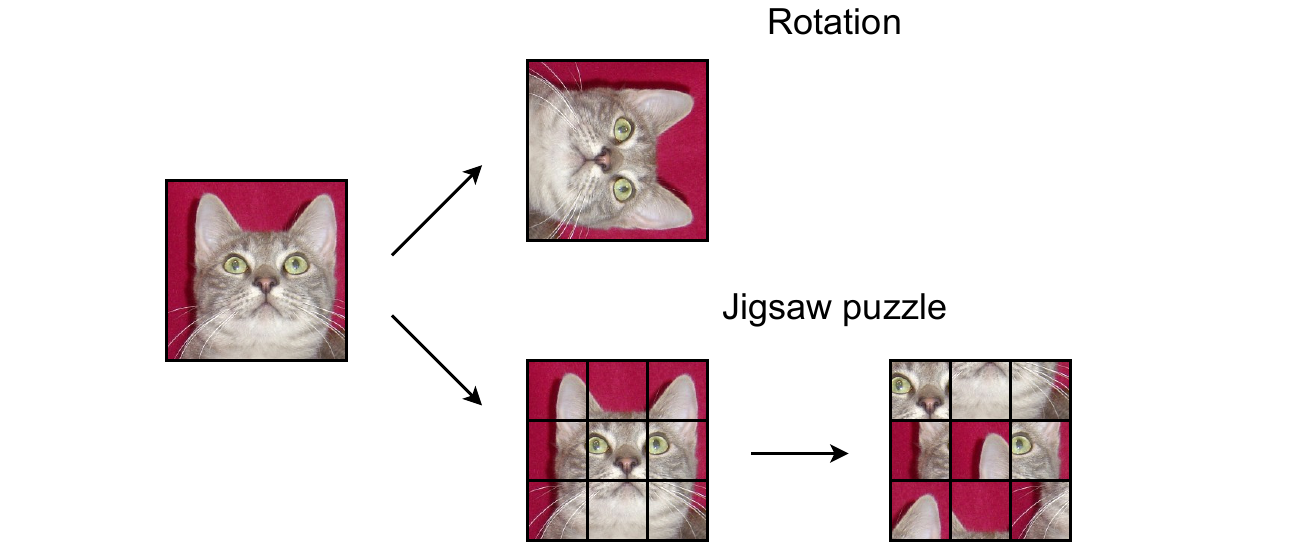}
    \caption{Classic spatial context-based methods}
    \label{fig:rotation_ssl}
  \end{subfigure}
  \hfill
  \begin{subfigure}{\linewidth}
    \centering
    \includegraphics[width=0.5\linewidth]{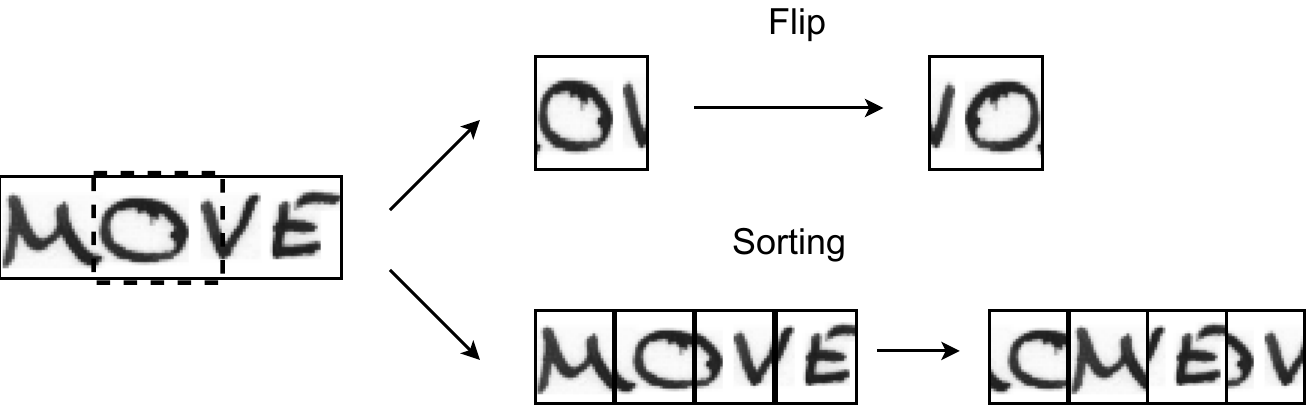}
    \caption{HTR spatial context-based methods}
    \label{fig:flip_ssl}
  \end{subfigure}
  \caption{(a) Existing spatial methods for SSL particularly designed for image classification characteristics. (b) We propose two new approaches adapted to the specific features of HTR: (i) ``Flip'' as a geometric transformation; (ii) ``Sorting'' as a puzzle transformation.}
  \label{fig:geometric_ssl}
\end{figure}

This work was motivated by the relevance of spatial information in HTR tasks~\cite{diaz2019perspective,luo2022siman} \review{together with the fact that, to our best knowledge, it has not been previously studied,}
and focuses on the adaptation of a reference set of such approaches to the HTR field. More precisely, we propose two alternatives that enable the use of these SSL strategies in HTR scenarios (see Fig.~\ref{fig:geometric_ssl}): (i) adapting the HTR task to the original SSL method, namely \textit{input adaptation}, and (ii) \review{proposing novel spatial context-based} SSL methods \review{devised} to the characteristics of the HTR task, namely \textit{method adaptation}. These proposals are comprehensively evaluated by considering several reference HTR corpora and are compared to the reference strategies followed in the field. The results obtained \review{show that the proposed SSL strategies provide state-of-the-art recognition rates compared to reference SSL approaches with less computational requirements, while maintaining conceptual simplicity.}  

The remainder of the work is structured as follows: Section~\ref{sec:related} contextualizes the current work within the related literature of the field, while Section~\ref{sec:method} introduces the strategies devised for spatial context-based SSL for HTR; Section~\ref{sec:experimental} presents the experimental set-up considered, and Section~\ref{sec:results} presents the results obtained; \review{Section~\ref{sec:discussion} provides an in-depth analysis and discussion of the strengths and limitations of the proposed methods;} finally, Section~\ref{sec:conclusions} concludes the work.

\section{Related work}\label{sec:related}

\subsection{Handwritten Text Recognition}
While HTR initially relied on Hidden Markov Models \cite{offline_tay_2001}, advances in Machine Learning favored the use of Recurrent Neural Networks (RNN) owing to their capacity to model sequential data. In particular, the use of bidirectional Long Short-Term Memory (LSTM) networks with Connectionist Temporal Classification (CTC)~\cite{connectionist_graves_2006} has been the state of the art in HTR for years \cite{VidaToselliRiosVilaCalvoZaragoza:PR:2023}. Despite this prevalence of CTC, attention-based (Att.) \review{autorregresive} encoder-decoder approaches \cite{neural_bahdanau_2015}, have recently gained popularity owing to their competitive results. For a detailed review of HTR approaches, see the work of \cite{evaluating_michael_2019}. 

As in other areas, there has been a growing interest in the use of more parallelizable architectures such as the Transformer \cite{TransformerVaswani} by adapting the work of \cite{Dosovitskiy2020AnII}. HTR has benefited from this adaptation, either in isolation with fully Transformer architectures \cite{trocr_li_2023} or in combination with other approaches \cite{dan_coquenet_2023}. Regardless of this progress, the need for large labelled corpora as a pre-training strategy in Transformer-based models has become noticeable \cite{trocr_li_2023}. This has led the SSL field to attract more interest in the HTR field owing to the usual scarcity of annotated data.

\subsection{Self-Supervised Text Recognition}

SSL represents a promising alternative for pre-training text recognition models, especially when addressing scenarios depicting scarcity of labelled data~\cite{alzubaidi2023survey}. In this context, the first work that tackled this issue was SeqCLR \cite{aberdam2021sequence}, which adapted the SimCLR contrastive learning method \cite{chen2020simple} to the sequential nature of text recognition. Note that, while this work is tested for both HTR and Scene Text Recognition (STR), subsequent SSL proposals have been specifically devised to address the nuances of each scenario. We now introduce some of the most relevant approaches in each case.

With regard to STR, PerSEC~\cite{liu2022perceiving} is a competitive example of a contrastive framework based on extracting low-level stroke and high-level semantic contextual spaces simultaneously via a hierarchical learning approach. SimAN~\cite{luo2022siman}, on the contrary, is framed within a generative strategy based on recovering the original format of an image crop by means of another non-overlapping crop of the same datum. This work highlights the close relationship between HTR and spatial context information. Other noticeable generative approaches rely on Mask Image Modelling (MIM) for the pre-training of full Transformer architectures, as occurs in the works by \cite{souibgui2023text} and \cite{yang2022reading}.

\review{In the context of HTR, the ChaCo~\cite{zhang2022chaco} and CMT-Co~\cite{zhang2022cmt} methods constitute representative examples of the contrastive learning paradigm. Both strategies take advantage of Momentum Contrast \cite{he2020momentum} to pre-train a visual feature extractor,} demonstrating that focusing solely on that part rather than the sequential can improve state-of-the-art performances in semi-supervised scenarios, which is also supported by the work of~\cite{luo2022siman}. \review{Regarding MIM, we find Text-DIAE \cite{souibgui2023text}, which is based on denoising autoencoders for Transformer architectures.}

\subsection{Spatial context-based methods}

Image data inherently encodes spatial information, which can be exploited to pre-train recognition models. We now present the two main families \review{of SSL} approaches that follow this principle highlighting the most important works related to them.

In terms of \emph{geometric transformations}, the work of~\cite{gidaris2018unsupervised} introduced the concept of predicting rotation alteration on image data as an SSL strategy. Its success in terms of both performance and simplicity led other works to further explore this principle in greater depth~\cite{novotny2018self,chen2019self,yamaguchi2021image}.

Since the relative position among patches of an image encompasses rich spatial and contextual information, there are several pretext tasks that exploit this premise, namely \emph{puzzle} strategies. Some of the most common approaches rely on predicting the relative positions of two patches of an image~\cite{doersch2015unsupervised} or recognising the order of a shuffled sequence of patches of an image~\cite{noroozi2016unsupervised,carlucci2019domain,yang2022fully}.

Despite being largely explored in classification frameworks, spatial context-based SSL is rarely exploited in image-based text recognition tasks. Nevertheless, the competitive results that are obtained by employing these strategies for classification, together with the strong relationship between spatial information and HTR---owing to stroke and font style~\cite{diaz2019perspective}---suggest that, if adequately adapted, text recognition models should benefit from pre-training methods of this nature. \review{Then, while previous efforts on SSL for HTR have focused on contrastive or image generation frameworks, this work explores the spatial context-based paradigm.} Note that, our work not only seeks to understand and unravel the existing spatial context SSL methods, but also presents innovative methods that specifically leverage and capitalize on this spatial context, thus providing new perspectives and advancements within HTR.

\section{Methodology}\label{sec:method}
\review{In this section, we introduce the general recognition framework considered and detail the spatial context-based SSL methods proposed in this work.}

\subsection{Encoder-decoder architecture}
\begin{figure}[tbh]
  \centering
  \includegraphics[width=0.8\linewidth]{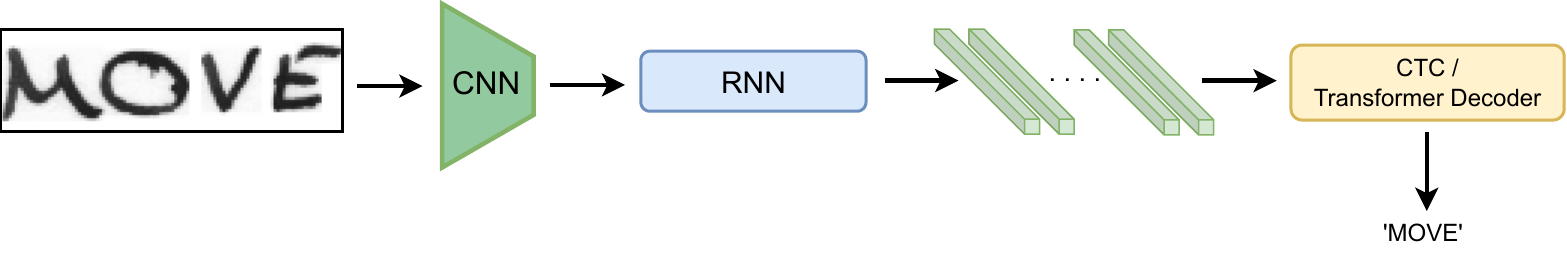}
  \caption{Encoder-decoder pipeline. The encoder is composed of a CNN that extracts the visual features for a given image. After a map-to-sequence operation, an RNN extracts the sequential features. The decoder, which may in this case be either CTC or a Transformer Decoder, transforms the representation into the output text.}
  \label{fig:encoder_decoder}
\end{figure}

A state-of-the-art encoder-decoder pipeline is adopted for the text recognition task, as shown in Fig.~\ref{fig:encoder_decoder}. The encoder is responsible for extracting the visual and sequential characteristics of an image, while the decoder is responsible for transforming the representation into the output text. 

\textbf{Encoder}. Given an image $I \in \mathbb{R}^{c \times H \times W}$ as input, where $c$ is the input channels, $H$ is the height and $W$ is the width, the first goes through a Convolutional Neural Network (CNN) $\text{cnn}: \mathbb{R}^{c\times H\times W}\rightarrow\mathbb{R}^{c^{\prime }\times H^{\prime}\times W^{\prime}}$, which is responsible for extracting visual features, where $c^{\prime}$ is the number of feature maps of size $H^{\prime}\times W^{\prime}$. A map-to-sequence operation $m:\mathbb{R}^{c^{\prime}\times H^{\prime}\times W^{\prime}}\rightarrow\mathbb{R}^{W^{\prime}\times\left(c^{\prime}\cdot H^{\prime}\right)}$ then reshapes the features into a sequence of frames. Finally, a RNN $\text{rnn}: \mathbb{R}^{W^{\prime}\times\left(c'\cdot H^{\prime}\right)}\rightarrow\mathbb{R}^{W^{\prime}\times T} $ is responsible for extracting sequential features, where $T$ is the output size that leads to the final representation $S = [s_{1}, s_{2}, ..., s_{W^{\prime}}]$. The encoder $f(\cdot)$ can, therefore, be expressed as $f(\cdot) = \text{rnn}(m(\text{cnn}(\cdot)))$

\textbf{Decoder}. Given the representation $S$, in this work two different text decoders are used: (i) a CTC decoder that separately decodes each frame of $S$, and (ii) a Transformer Decoder (TD), which is responsible for autoregressively producing the text characters \review{from the representation S} and the previously generated output.

\subsection{SSL pipeline}
\begin{figure*}[tbh]
  \centering
  \includegraphics[width=0.8\linewidth]{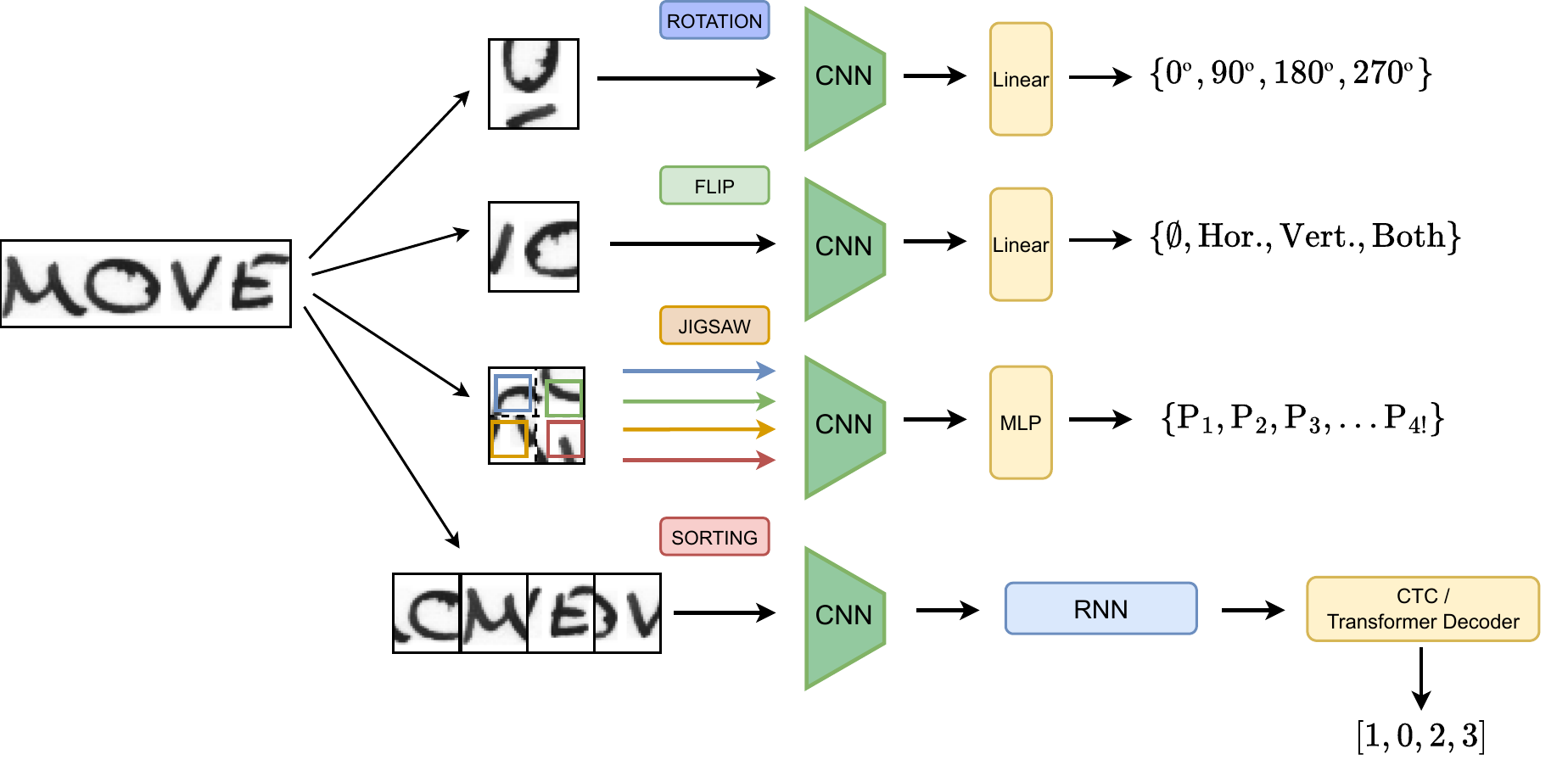}
  \caption{Pre-training pipeline for the different pretext tasks proposed in the work. An image is transformed according to the method and the subsequent layers handle the task. Note that ``Rotation'', ``Flip'' and ``Jigsaw'' predict a single class from the sets presented, while ``Sorting'' predicts a sequence.}
  \label{fig:SSL_pipeline}
\end{figure*}

This section presents the proposed adaptations \review{for} the main existing spatial context-based pretext tasks to HTR scenarios (see Fig.~\ref{fig:SSL_pipeline}). \review{Note that for each spatial SSL family---i.e. geometric and puzzle---we have chosen the most representative pretext task. Then,} for each \review{selected} method, we present two possible means of enabling the use of these techniques in HTR scenarios: (i) adapting the data to a classification-oriented framework (\textit{input adaptation}), and (ii) modifying the underlying nature of the pretext task to allow its used in HTR-based data (\textit{method adaptation}).

\subsubsection{Geometric transformations}
As stated previously, one of the most relevant strategies in terms of geometric transformations is the prediction of rotations when applied to an image~\cite{gidaris2018unsupervised}. While a transformation of this nature can be applied directly in classification scenarios, the typology of the input in HTR data makes its direct use complicated, since: i) text images are rarely squared, signifying that most rotations alter the dimensionality of the axes; and ii) applying a rotation to such elements drastically changes the semantic meaning of the axes (characteristics and sequential order). \review{Therefore, we explore} two possible adaptations of this technique to HTR scenarios:

\begin{itemize}
    \item \textbf{Rotation (input adaptation)}. Given that images in HTR rarely depict the same width, a fixed-size crop---for simplicity, we consider that a square crop of dimensions $H\times H$---of the input data must be extracted to enable the use of this pretext task. This pre-training process can be formally described as follows: given an image $I$, a squared crop is made, obtaining $I_{square}$, after which a rotation is uniformly chosen from the set \{0º, 90º, 180º, 270º\}, retrieving image $I_{r}$. Finally, the CNN $cnn(\cdot)$ will, together with a linear classification layer $l(\cdot)$, be responsible for solving the pretext task.

    \item \textbf{Flip (method adaptation)}. An alternative to the aforementioned case consists of applying a flip transformation rather than the rotation transformation. The advantage of this for HTR is that it does not modify the sequential dimension of the data. Moreover, the unique asymmetry of writing, which lacks symmetry both vertically and horizontally, requires a model to have the ability to understand letter shapes, carry out recognition and discern whether a flip has been applied. Formally, given an image $I$, a crop is made in order to give the image $I_{crop}$. This element is then transformed using a flipping process that is randomly selected from the set \{$\phi$, Horizontal (Hor.), Vertical (Vert.), Hor. and Vert. (Both)\}. Finally, the CNN $cnn(\cdot)$, together with a linear classification layer $l(\cdot)$, predicts the transformation that has been applied.
\end{itemize}

\subsubsection{Puzzle transformation}
One of the most prominent methods in this family is the so-called \emph{jigsaw puzzle}, which consists of rearranging a set of jumbled patches from the same image by predicting the index of the correct permutation~\cite{noroozi2016unsupervised}. Since this proposal was conceived for classification tasks in which images depict the same dimensions, we now present the proposals to enable its use in HTR images with different width dimensions.

\begin{itemize}
    \item \textbf{Jigsaw puzzle (input adaptation)}. The original method is maintained by extracting a square crop from the HTR image, and this is later processed by employing the jigsaw puzzle strategy. Formally, for a given image $I$, a square crop of $H\times H$ dimensions is extracted from the image, obtaining $I_{square}$; the image is initially reshaped to obtain $I_{square}^{\prime} \in \mathbb{R}^{c\times 150\times 150}$ and then divided into 4 shuffled patches of dimensions $75 \times 75$; these patches are further cropped into elements of size $64\times64$, \emph{i.e.}, $P = [p_1, p_2, p_3, p_4]$, where $p_i \in \mathbb{R}^{1\times 64\times 64}$; each patch is subsequently processed by the CNN feature extractor, whose results are concatenated and fed into a multi-layer perceptron that estimates the permutation performed \review{among the $24$ ($4!$) possibilities}.
    \item \textbf{Sorting (method adaptation)}. The proposed method adaptation consists of rearranging a fixed-size non-overlapping number of patches---experimental parameter---into which the HTR image is divided. Formally, given an input image $I$ and a number of patches $M$---in this case, $M \in \{2, 3, 4\}$---, a list of patches $P = [p_1, ..., p_M]$, where $p_i \in \mathbb{R}^{1\times H \times \lfloor W/M\rfloor}$, is obtained. $P$ is then randomly unsorted and rejoined to form the image $I_{shuffled}$. Finally, the encoder $f(\cdot)$ predicts the correct \review{index} order of the patches using an auxiliary decoder.
    
\end{itemize}

\section{Experimental setup}\label{sec:experimental}
This section presents the datasets and the metric considered in order to validate the proposals presented, along with the information regarding the neural architectures used and their implementation details.

\subsection{Datasets}
The proposed methods are assessed using three reference public datasets for HTR: the IAM Handwritten Database~\cite{marti2002iam}, the CVL database (CVL)~\cite{kleber2013cvl}, and RIMES~\cite{grosicki2024rimes}. Table~\ref{tab:corpora} provides a summary of the characteristics of the corpora at a word-level. For the pre-training, the SSL methods described above use only samples of the training split. We consider the Aachen split\footnote{\url{https://www.openslr.org/56/}} for the IAM set. Since CVL does not contain an official validation split, we use 5\% of the training data for validation. \review{In RIMES, we ensure text-letter independence (more details in Supplementary materials)}. In all cases, we transform the input images to grayscale and normalize them to $H=64$, maintaining the aspect ratio.

\begin{table}[htb]
\caption{Overview of the corpora used in this work, depicting their characteristics (number of samples, size of the alphabet, number of writers, time period, and language).}
\label{tab:corpora}
\centering

    \setlength{\tabcolsep}{3pt}
    \begin{tabular}{l c c c l}
    \toprule[1pt]
        & \textbf{Images} & \textbf{Alphabet}& \textbf{Writers} & \textbf{Language} \\
        \cmidrule(lr){2-5} 
        \textbf{IAM} & 79\,275 & 78& 657 & Eng \\
        \textbf{RIMES} & 228\,358 & 95 & 1\,300 & French \\
        \textbf{CVL} & 99\,904 & 53& 310 & Eng, Ger\\
    \bottomrule[1pt]
\end{tabular}
\end{table}

\review{For comparative purposes with the related literature \cite{zhang2022chaco, souibgui2023text}, performance was assessed in terms of Word Accuracy (WAcc). This figure of merit is defined as the percentage of predicted words that exactly matches the corresponding ground truth.}

\subsection{Neural architectures}
The details of the considered encoder-decoder recognition scheme are presented below:

\begin{itemize}
    \item \textbf{Encoder}: \review{We resorted to the ResNet-29 model used in the works of~\cite{cheng2017focusing} and \cite{aberdam2021sequence} to provide feasible comparison, but with slight modifications to adapt it to our input characteristics.} \review{More details are provided in the Supplementary material.}

    \item \textbf{Recurrent feature extractor}: The sequential nature of the HTR data was modelled by considering two stacked bidirectional LSTMs with a hidden size of $T=256$.

    \item \textbf{Transformer decoder}: We followed the vanilla decoder used by~\cite{TransformerVaswani}, fixing a single decoding layer with 8 attention heads and a hidden size of 512. 

\end{itemize}

With regard to the training details, a maximum number of 1000 epochs with a patience of 50 on the validation loss has been considered for both the pre-training and the downstream stages. Note that, in the latter case, this patience is increased to 200 epochs when a reduced part of the dataset is used. Only the training data is used for the self-supervised pre-training. Data augmentation is also contemplated in both stages, the details of which are provided in the Supplementary material. The weights of the model are optimized using the Adam method~\cite{adamKingmaB14}, fixing the learning rate values to $3\cdot 10^{-4}$ and $10^{-4}$ for the pre-training and fine-tuning phases, respectively. 

\section{Results}\label{sec:results}
The results of our experiments are presented below, and are divided into two main sections. In the first, we analyze the quality of the representations learned for HTR by the different methods. In the second, we compare these methods in a transfer learning scenario, considering cases with both limited and abundant data for fine tuning. \review{In all cases, we compare against state-of-the-art SSL methods for HTR}.


\subsection{Quality evaluation}
The evaluation of SSL methods often hinges on the quality of the representations learned for the task, without the aid of any auxiliary training. This quality directly represents the ability of the model to learn broad and general representations. To assess this aspect in the context of HTR, we freeze all representation layers after pre-training to then fine-tune \review{the rest} of the model using all the training data. By freezing the representation layers, we ensure that the figures reporting recognition accuracy can be attributed to the quality of the representations themselves, rather than enhancements in the architecture of the model or the fine-tuning process as a whole.

Table~\ref{tab:resuls_qualityEvaluation} provides details of the quality assessment \review{for each method and dataset. First,} focusing on our two primary SSL blocks---Geometric and Puzzle---it will be noted that \review{the Geometric-based methods show a clear superiority against the Puzzle-based ones while being the last still competitive in several cases, thus demonstrating the utility of spatial information in HTR tasks. Concerning the best-performing options, ChaCo seems to be the most effective in general terms, although the Geometric strategies are close in performance. In this regard, the average differences for each case between the best-performing method among the Baselines and the top Geometric method are 6.2 and 2.8 WAcc points for CTC and autoregressive decoders, respectively.}

\begin{table}[!ht]
    \caption{WAcc (\%) obtained in the representation quality evaluation for our proposals and state-of-the-art methods. Bold valued indicate best value per dataset and decoding. Underlined values denote best value per dataset and decoding among the proposals. $\dag$ and $\ddag$ indicate the methods that use Att. or TD in autorregresive column, respectively.}
    \label{tab:resuls_qualityEvaluation}
    \centering
    \setlength{\tabcolsep}{3pt}
    \renewcommand{\arraystretch}{1.1}
    \begin{tabular}{llccccccc}
    \toprule[1pt]
    \multicolumn{2}{l}{\multirow{2}{*}{\textbf{Method}}} & \multicolumn{3}{c}{\textbf{CTC}} &  & \multicolumn{3}{c}{\textbf{Autorregresive}} \\
    \cmidrule{3-5}\cmidrule{7-9}
    \multicolumn{2}{l}{} & IAM & CVL & RIMES &  & IAM & CVL & RIMES \\
    \cmidrule{1-9}
    \multicolumn{2}{l}{\textit{Baselines}} &  &  &  &  &  &  &  \\
     & SeqCLR\dag & 39.7 & 66.7 & 63.8 &  & 51.9 & 74.5 & 79.5 \\
     & ChaCo\dag & \textbf{70.0} & \textbf{75.6} & 79.4 &  & \textbf{72.9} & 77.6 & \textbf{84.9} \\
     & CMT-Co\dag & 53.1 & 72.1 & 66.0 &  & 58.2 & 74.7 & 74.7 \\
     & Text-DIAE\ddag & - & - & - &  & 71.0 & \textbf{78.1} & - \\
     \cmidrule{1-9}
    \multicolumn{2}{l}{\textit{Geometric}} &  &  &  &  &  &  &  \\
     & Rotation\ddag & 65.2 & 54.9 & \underline{\textbf{85.0}} &  & \underline{67.6} & \underline{77.2} & 82.4 \\
     & Flip\ddag & \underline{65.9} & \underline{55.4} & 82.0 &  & 50.1 & 67.3 & 57.1 \\
    \multicolumn{2}{l}{\textit{Puzzle}} &  &  &  &  &  &  &  \\
     & Jigsaw\ddag & 36.4 & 18.1 & 53.9 &  & 46.2 & 46.5 & 64.7 \\
     & Sorting\ddag & 55.0 & 26.4 & 66.1 &  & 63.3 & 52.7 & \underline{84.3} \\
     \bottomrule[1pt]
    \end{tabular}
\end{table}

\subsection{Transfer learning}
We now explore a transfer learning scenario to evaluate the adaptability and efficiency of the pre-trained SSL methods. The essence of this experiment is to observe how well these methods can leverage a certain amount of labelled data---5\%, 10\%, or 100\% of the dataset---to fine tune the entire network starting from the representations learned \review{during pre-training}. This fine-tuning process involves all layers, which contrasts with the previous experiment in which only the final layers were updated. We aim to discern the ability of these methods to obtain useful representations in HTR as a downstream task and, ultimately, to achieve the highest possible recognition accuracy. \review{Note that, the same random samples are considered for each method when using a reduced part of a dataset.}

Table~\ref{tab:resuls_transferLearning} presents the results of this experiment \review{reporting the WAcc for all datasets and methods. For the spatial context-based SSL strategies, the Geometric approaches, particularly the ``Flip" proposal, demonstrate notable superiority. Specifically, this method exhibits a notable advantage over state-of-the-art methods when using a reduced portion of the training data in the IAM and RIMES datasets. For instance, in the IAM dataset with only 10\% of the data, ``Flip" outperforms the previously best-performing method in each scenario by 11.4 and 11.1 points for the CTC and TD decoders, respectively. Although the CVL dataset does not show this superiority in data-reduced scenarios, ``Flip" still surpasses the previous best method for each case when using the full dataset by 7.3 and 4 points for the CTC and TD decoders, respectively. These results highlight the method's ability to achieve substantial progress in state-of-the-art performance across different scenarios.}

\review{On the other hand, it is also worth highlighting the ``Rotation" method, which, while slightly trailing ``Flip" in performance, still achieves remarkable improvements over the previous state-of-the-art methods in many scenarios. Regarding the Puzzle-based methods, while they remain competitive, they do not demonstrate better performance over either the state-of-the-art or the Geometric-based methods.}

\begin{table*}[!ht]
    \caption{WAcc (\%) metric across all datasets and methods for the transfer learning scheme, considering different percentages of labelled data. Bold values denote the best results obtained per percentage of labelled data. Underlined values indicate best value per dataset and among the proposals. $\dag$ and $\ddag$ indicate the methods that use Att. or TD in the autorregresive table, respectively.}
    \label{tab:resuls_transferLearning}
    \centering
    \renewcommand{\arraystretch}{1}

    \begin{subtable}{\linewidth}
        \centering
        \begin{tabular}{llccccccccccc}
        \toprule[1pt]
        \multicolumn{2}{l}{\multirow{2}{*}{\textbf{Method}}} & \multicolumn{3}{c}{\textbf{IAM}} &  & \multicolumn{3}{c}{\textbf{CVL}} &  & \multicolumn{3}{c}{\textbf{RIMES}} \\ \cline{3-5} \cline{7-9} \cline{11-13} 
        \multicolumn{2}{l}{} & 5 & 10 & 100 &  & 5 & 10 & 100 &  & 5 & 10 & 100 \\ \hline
        \multicolumn{2}{l}{\textit{Baselines}} &  &  &  &  &  &  &  &  &  &  &  \\
         & SeqCLR & 31.2 & 44.9 & 76.7 &  & 66.0 & 71.0 & 77.0 &  & 61.8 & 71.9 & 90.1 \\
         & ChaCo & 44.3 & 53.1 & 79.8 &  & 65.6 & 70.3 & 78.1 &  & 61.3 & 73.0 & 89.2 \\
         & CMT-Co & 47.7 & 56.0 & \textbf{80.1} &  & \textbf{71.9} & \textbf{74.5} & 78.2 &  & 67.1 & 75.4 & 90.1 \\
         & PerSec-CNN & - & - & 77.9 &  & - & - & 78.1 &  & - & - & - \\ \hline
        \multicolumn{2}{l}{\textit{Geometric}} &  &  &  &  &  &  &  &  &  &  &  \\
         & Rotation & 60.2 & 66.3 & 79.0 &  & 39.0 & 52.5 & 85.3 &  & 79.7 & \underline{\textbf{84.4}} & \underline{\textbf{91.4}} \\
         & Flip & \underline{\textbf{61.7}} & \underline{\textbf{67.4}} & \underline{79.1} &  & \underline{40.6} & \underline{54.6} & \underline{\textbf{85.5}} &  & \underline{\textbf{79.8}} & 84.2 & 91.1 \\
        \multicolumn{2}{l}{\textit{Puzzle}} &  &  &  &  &  &  &  &  &  &  &  \\
         & Jigsaw & 54.5 & 64.0 & 77.7 &  & 28.7 & 42.1 & 83.8 &  & 77.6 & 81.0 & 90.8 \\
         & Sorting & 56.4 & 64.7 & 79.0 &  & 31.8 & 34.7 & 78.5 &  & 74.1 & 71.5 & 90.0 \\
         \bottomrule[1pt]
        \end{tabular}
        \caption{CTC decoder.}
        \label{tab:resuls_transferLearning_ctc}
    \end{subtable}

    \begin{subtable}{\linewidth}
        \centering
        \begin{tabular}{llccccccccccc}
        \toprule[1pt]
        \multicolumn{2}{l}{\multirow{2}{*}{\textbf{Method}}} & \multicolumn{3}{c}{\textbf{IAM}} &  & \multicolumn{3}{c}{\textbf{CVL}} &  & \multicolumn{3}{c}{\textbf{RIMES}} \\ \cline{3-5} \cline{7-9} \cline{11-13} 
        \multicolumn{2}{l}{} & 5 & 10 & 100 &  & 5 & 10 & 100 &  & 5 & 10 & 100 \\ \hline
        \multicolumn{2}{l}{\textit{Baselines}} &  &  &  &  &  &  &  &  &  &  &  \\
         & SeqCLR\dag & 40.3 & 52.3 & 79.9 &  & 73.1 & 74.8 & 77.8 &  & 70.9 & 77.0 & \textbf{92.5} \\
         & ChaCo\dag & 46.7 & 55.5 & 81.4 &  & 68.9 & 74.0 & 78.2 &  & 68.4 & 77.0 & 90.6 \\
         & CMT-Co\dag & 50.4 & 55.8 & \textbf{81.9} &  & \textbf{73.6} & \textbf{76.2} & 78.7 &  & 68.8 & 77.6 & 91.2 \\
         & PerSec-CNN\dag & - & - & 80.8 &  & - & - & 80.2 &  & - & - & - \\
         & Text-DIAE\ddag & 49.6 & 58.7 & 80.0 &  & 47.9 & 68.5 & 87.3 &  & - & - & - \\ \hline
        \multicolumn{2}{l}{\textit{Geometric}} &  &  &  &  &  &  &  &  &  &  &  \\
         & Rotation\ddag & \underline{\textbf{62.9}} & 68.1 & 79.6 &  & \underline{53.5} & \underline{66.5} & 91.2 &  & 80.3 & \underline{\textbf{84.8}} & 91.7 \\
         & Flip\ddag & 62.5 & \underline{\textbf{69.8}} & \underline{80.8} &  & 52.1 & 66.3 & \underline{\textbf{91.3}} &  & 79.0 & 83.9 & 89.7 \\
        \multicolumn{2}{l}{\textit{Puzzle}} &  &  &  &  &  &  &  &  &  &  &  \\
         & Jigsaw\ddag & 58.5 & 68.2 & 80.7 &  & 43.5 & 63.3 & 90.1 &  & 78.9 & 83.2 & \underline{92.0} \\
         & Sorting\ddag & 59.2 & 69.2 & \underline{80.8} &  & 43.4 & 54.2 & 86.2 &  & \underline{\textbf{80.4}} & 83.9 & 91.7 \\
         \bottomrule[1pt]
        \end{tabular}
        \caption{Autorregresive decoder.}
        \label{tab:resuls_transferLearning_autorregresive}
    \end{subtable}

\end{table*}

\section{\review{Discussion}} \label{sec:discussion}


\review{In this section, we delve deeper into the analysis and interpretation of the results. For this purpose, we also visualize the feature extraction of each spatial context-based SSL method, compare efficiency of the different neural architectures, and examine the strengths and limitations of the proposed methods.}

\review{\subsection{Comparison and feature visualization}}

\begin{figure}[t]
  \centering
  \begin{subfigure}{0.49\linewidth}
    \centering
    \includegraphics[width=0.5\linewidth]{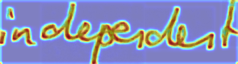}
    \caption{Rotation}
    \label{fig:image_cam_rotation}
  \end{subfigure}
  \hfill
  \begin{subfigure}{0.49\linewidth}
    \centering
    \includegraphics[width=0.5\linewidth]{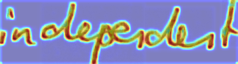}
    \caption{Flip}
    \label{fig:image_cam_flip}
  \end{subfigure}
  \begin{subfigure}{0.49\linewidth}
    \centering
    \includegraphics[width=0.5\linewidth]{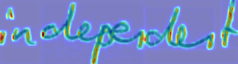}
    \caption{Jigsaw}
    \label{fig:image_cam_jigsaw}
  \end{subfigure}
  \hfill
  \begin{subfigure}{0.49\linewidth}
    \centering
    \includegraphics[width=0.5\linewidth]{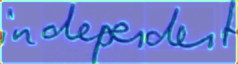}
    \caption{Sorting}
    \label{fig:image_cam_sorting}
  \end{subfigure}
  \caption{Heat map highlighting the regions targeted by the encoder pre-trained with the different spatial context-based SSL methods.}
  \label{fig:cam}
\end{figure}

Considering the results reported, it is worth noting a rather counterintuitive phenomenon: the only method that leverages the sequential nature of the whole word image (``Sorting'') does not lead the best representations. This outcome is surprising, as one would expect this pretext task to provide a comprehensive view of the input, closely mirroring what the network experiences during the actual HTR task. In contrast, partial images (crops) yield more informative representations for HTR, at least from the point of view of pure pre-training. \review{This argument is supported by Fig. \ref{fig:cam}, which visualizes the feature extraction of the proposed methods as a heat map, highlighting that ``Sorting" produces less effective visual features than the Geometric approaches. In addition, note that the ``Jigsaw" method, which focuses at sub-character level (just partial views of characters), also shows a clear worse attention to the stroke patterns}. This reinforces some of the conclusions drawn in the literature, which suggest that focusing on local patterns such as single characters or strokes favours the learning process~\cite{luo2022siman,zhang2022chaco}.

It may be also observed that \review{in our approaches,} the TD setting systematically makes better use of the representations learned through SSL. This trend is evident, as all the TD settings outperform their counterparts that employ CTC. The representations learned would, therefore, appear to reflect an intrinsic compatibility with the TD approach for the decoding of character sequences. Furthermore, the order of methods according to their performance remains consistent regardless of the decoder utilized, whether CTC or TD. This consistency suggests that the quality of the representations is inherently sound across different decoders, a valuable trait for the development of robust HTR systems.

At last, although many of our proposals reflect robust behavior that equal on the state of the art, the ``Flip'' method is positioned as the best-performing alternative in general terms. \review{This method not only surpasses other spatial context-based techniques but also outperforms prior SSL methods in the semi-supervised scenario. In particular, ``Flip" exhibits clear superiority over the ``Rotation" strategy when using CTC decoding. We believe that this advantage arises from Flip's preservation of the horizontal orientation of text, which aligns well with CTC's horizontal decoding of feature representations.} This reinforces the value of developing and implementing strategies that are directly aligned to the unique challenges and characteristics of HTR task.

\review{\subsection{Computational architectural requirements}}

\review{While CRNN architectures with CTC decoding have dominated the state of the art for years, transformer architectures have recently demonstrated exceptional effectiveness in HTR tasks mainly due to their low inductive bias and language model ability. Nevertheless, this advantage comes at the expense of increased computational complexity and resource demands \cite{liu2022perceiving}. Given these trade-offs, it is essential to assess the efficiency and practicality of the chosen architecture. Therefore, this section analyzes the computational requirements of the architectures considered in the work and compares them with the state-of-the-art SSL methods.}

\begin{table}[!ht]
    \caption{Table showing the input size, GFLOPs, and number of parameters for each method. Symbol * denotes that the number of columns is dynamically adjusted to preserve the image aspect ratio.}
    \label{tab:computational_comparison}
    \centering
    \setlength{\tabcolsep}{4pt}
    \renewcommand{\arraystretch}{1.1}
    \begin{tabular}{llccc}
    \toprule[1pt]
    \textbf{Architecture} & \multicolumn{1}{c}{} & \textbf{Input size} & \textbf{GFLOPs} & \textbf{Params} \\ \cline{1-1} \cline{3-5} 
    CTC (ours) & \multicolumn{1}{c}{} & {[}64,*{]} & 20.4 & 48.0M \\
    TD (ours) & \multicolumn{1}{c}{} & {[}64,*{]} & 20.7 & 50.7M \\
    \cite{cheng2017focusing} CTC &  & {[}32, 100{]} & 9.7 & 48.5M \\
    \cite{cheng2017focusing} Att. &  & {[}32, 100{]} & 24.5 & 49.5M \\
    Text-DIAE &  & {[}64, 256{]} & 27.6 & 68.8M \\
    \bottomrule[1pt]
    \end{tabular}
\end{table}

\review{Table \ref{tab:computational_comparison} presents the Floating Point Operations (FLOPs) and the number of parameters for the architectures used by the different methods for which we were able to replicate the architecture. Focusing first on the architectures employed in this work, as expected, the one incorporating a TD decoder exhibits higher FLOP consumption. However, since the decoder consists of only one layer, the additional computational cost remains minimal---just 0.3 GFLOPs. This makes it a highly favorable trade-off, offering a modest increase in resource usage and parameters while delivering a substantial improvement in performance. In contrast, Text-DIAE, a Transformer-based method, emerges as the most resource-intensive architecture among the evaluated but it does not deliver the best results as shown in Table \ref{tab:resuls_transferLearning_autorregresive}, underscoring the fact that simpler architectures and methods can surpass these more complex approaches in performance. This highlights the importance of balancing architectural complexity with efficiency and effectiveness in HTR systems.}

\review{On the other hand, when using CTC, our models exhibit more FLOP consumption compared to other CRNN-based methods (note that ChaCo, SeqCLR and CMT-Co share the architecture of \cite{cheng2017focusing}) despite having a similar number of parameters. This is because the input image size in our case is, typically, larger. However, this issue is notable mitigated by the inherent computational simplicity of the methods, as we will discuss in next section.}

\review{\subsection{Strengths and limitations}}

\review{In recent years, contrastive learning or dual-branch architectures has significantly advanced capabilities in image classification. However, its direct application to HTR is not straightforward. For example, aggressive data augmentation can cause misalignment between positive and negative pairs, leading to suboptimal representations \cite{aberdam2021sequence}. In contrast, the spatial context-based SSL methods explored in this work avoid such issues. By focusing solely on stroke patterns within the processed images, these methods can leverage far more aggressive data augmentations, such as strong geometric transformations or croppings. This approach enables the network to encounter a wider variety of data, fostering a deeper understanding of their intrinsic characteristics. Additionally, the simplicity of the presented pretext tasks---\emph{e.g.}, classifying an image crop in terms of its rotation angle---reduces computational training costs compared to dual-branch architectures or generative approaches while achieving similar performance rates.}

\review{Concerning the limitations, it has also been proven that dual-branch architectures or generative methods show a higher abstraction capacity in the representations by encapsulating more semantic information. In this regard, since the presented proposals exclusively focus on the visual information extracted from the stroke and font style, its applicability to other off-the-shelf tasks---\emph{e.g.}, text segmentation, writer retrieval, image enhancement, multilingual or historical manuscripts---remains uncertain. Future works could explore the generalizability of the different SSL methods.}

\section{Conclusions}\label{sec:conclusions}
In this paper, we study spatial context-based SSL methods originally devised for classification tasks and adapt them to the HTR field. We propose two possible adaptation frameworks for this purpose: (i) adjusting the data in order to use the original SSL methods (input adaptation), and (ii) adapting these SSL approaches to the sequential nature of HTR (method adaptation). The results obtained when considering three benchmark HTR datasets---IAM, CVL, and RIMES---show that these proposals achieve state-of-the-art performance in various scenarios, particularly in semi-supervised settings. Moreover, \review{among the proposals, it should be noted that one specifically designed for HTR (``Flip'') seems to be specially effective}. Also note that our results are aligned with those of previous SSL literature on text recognition~\cite{luo2022siman}, which argues that there is a close relationship between HTR and spatial context owing to stroke and font style information. Then, the final objective of this study is to provide practical guidelines to take advantage of a family of SSL techniques in HTR context, thus encouraging the study and application of different SSL frameworks in this field.

\bibliographystyle{splncs04}
\bibliography{main}  

\begin{thebibliography}{10}
\providecommand{\url}[1]{\texttt{#1}}
\providecommand{\urlprefix}{URL }
\providecommand{\doi}[1]{https://doi.org/#1}

\bibitem{aberdam2021sequence}
Aberdam, A., Litman, R., Tsiper, S., Anschel, O., Slossberg, R., Mazor, S., Manmatha, R., Perona, P.: Sequence-to-sequence contrastive learning for text recognition. In: Proceedings of the IEEE/CVF CVPR. pp. 15302--15312 (2021)

\bibitem{alzubaidi2023survey}
Alzubaidi, L., Bai, J., Al-Sabaawi, A., Santamar{\'\i}a, J., Albahri, A., Al-dabbagh, B.S.N., Fadhel, M.A., Manoufali, M., Zhang, J., Al-Timemy, A.H., et~al.: A survey on deep learning tools dealing with data scarcity: definitions, challenges, solutions, tips, and applications. Journal of Big Data  \textbf{10}(1), ~46 (2023)

\bibitem{neural_bahdanau_2015}
Bahdanau, D., Cho, K., Bengio, Y.: Neural machine translation by jointly learning to align and translate  (2015)

\bibitem{balestriero2023cookbook}
Balestriero, R., Ibrahim, M., Sobal, V., Morcos, A., Shekhar, S., Goldstein, T., Bordes, F., Bardes, A., Mialon, G., Tian, Y., et~al.: A cookbook of self-supervised learning. arXiv preprint arXiv:2304.12210  (2023)

\bibitem{BezerraZanchettinToseeliPirlo:Book:2017}
Bezerra, B.L.D., Zanchettin, C., Toselli, A.H., Pirlo, G.: Handwriting: recognition, development and analysis. Nova Science Publishers, Inc. (2017)

\bibitem{carlucci2019domain}
Carlucci, F.M., D'Innocente, A., Bucci, S., Caputo, B., Tommasi, T.: Domain generalization by solving jigsaw puzzles. In: Proceedings of the IEEE/CVF CVPR. pp. 2229--2238 (2019)

\bibitem{chen2020simple}
Chen, T., Kornblith, S., Norouzi, M., Hinton, G.: A simple framework for contrastive learning of visual representations. In: International conference on machine learning. pp. 1597--1607. PMLR (2020)

\bibitem{chen2019self}
Chen, T., Zhai, X., Ritter, M., Lucic, M., Houlsby, N.: Self-supervised gans via auxiliary rotation loss. In: Proceedings of the IEEE/CVF CVPR. pp. 12154--12163 (2019)

\bibitem{cheng2017focusing}
Cheng, Z., Bai, F., Xu, Y., Zheng, G., Pu, S., Zhou, S.: Focusing attention: Towards accurate text recognition in natural images. In: Proceedings of the IEEE ICCV. pp. 5076--5084 (2017)

\bibitem{dan_coquenet_2023}
Coquenet, D., Chatelain, C., Paquet, T.: Dan: a segmentation-free document attention network for handwritten document recognition  (2023)

\bibitem{diaz2019perspective}
Diaz, M., Ferrer, M.A., Impedovo, D., Malik, M.I., Pirlo, G., Plamondon, R.: A perspective analysis of handwritten signature technology. Acm Computing Surveys  \textbf{51}(6),  1--39 (2019)

\bibitem{doersch2015unsupervised}
Doersch, C., Gupta, A., Efros, A.A.: Unsupervised visual representation learning by context prediction. In: Proceedings of the IEEE ICCV. pp. 1422--1430 (2015)

\bibitem{Dosovitskiy2020AnII}
Dosovitskiy, A., Beyer, L., Kolesnikov, A., Weissenborn, D., Zhai, X., Unterthiner, T., Dehghani, M., Minderer, M., Heigold, G., Gelly, S., Uszkoreit, J., Houlsby, N.: An image is worth 16x16 words: Transformers for image recognition at scale. ArXiv  \textbf{abs/2010.11929} (2020)

\bibitem{gidaris2018unsupervised}
Gidaris, S., Singh, P., Komodakis, N.: Unsupervised representation learning by predicting image rotations. arXiv:1803.07728  (2018)

\bibitem{connectionist_graves_2006}
Graves, A., Fernández, S., Gomez, F.J., Schmidhuber, J.: Connectionist temporal classification: labelling unsegmented sequence data with recurrent neural networks. ICML  (2006)

\bibitem{grosicki2024rimes}
Grosicki, E., Carré, M., Geoffrois, E., Augustin, E., Preteux, F., Messina, R.: Rimes, complete [data set]  (2024)

\bibitem{he2020momentum}
He, K., Fan, H., Wu, Y., Xie, S., Girshick, R.: Momentum contrast for unsupervised visual representation learning. In: Proceedings of the IEEE/CVF CVPR. pp. 9729--9738 (2020)

\bibitem{jaiswal2020survey}
Jaiswal, A., Babu, A.R., Zadeh, M.Z., Banerjee, D., Makedon, F.: A survey on contrastive self-supervised learning. Technologies  \textbf{9}(1), ~2 (2020)

\bibitem{jing2020self}
Jing, L., Tian, Y.: Self-supervised visual feature learning with deep neural networks: A survey. IEEE Trans Pattern Anal Mach Intell  \textbf{43}(11),  4037--4058 (2020)

\bibitem{jung2020imgaug}
Jung, A.B., Wada, K., Crall, J., Tanaka, S., Graving, J., Reinders, C., Yadav, S., Banerjee, J., Vecsei, G., Kraft, A., et~al.: imgaug. GitHub: San Francisco, CA, USA  (2020)

\bibitem{adamKingmaB14}
Kingma, D.P., Ba, J.: Adam: {A} method for stochastic optimization. In: Bengio, Y., LeCun, Y. (eds.) 3rd ICLR (2015)

\bibitem{kleber2013cvl}
Kleber, F., Fiel, S., Diem, M., Sablatnig, R.: Cvl-database: An off-line database for writer retrieval, writer identification and word spotting. In: 2013 12th ICDAR. pp. 560--564. IEEE (2013)

\bibitem{trocr_li_2023}
Li, M., Lv, T., Chen, J., Cui, L., Lu, Y., Florencio, D., Zhang, C., Li, Z., Wei, F.: Trocr: Transformer-based optical character recognition with pre-trained models. Proceedings of the ... AAAI Conference on Artificial Intelligence  (2023)

\bibitem{liu2022perceiving}
Liu, H., Wang, B., Bao, Z., Xue, M., Kang, S., Jiang, D., Liu, Y., Ren, B.: Perceiving stroke-semantic context: Hierarchical contrastive learning for robust scene text recognition. In: Proceedings of the AAAI Conference on Artificial Intelligence. vol.~36, pp. 1702--1710 (2022)

\bibitem{liu2021self}
Liu, X., Zhang, F., Hou, Z., Mian, L., Wang, Z., Zhang, J., Tang, J.: Self-supervised learning: Generative or contrastive. IEEE Trans Knowl Data Eng  \textbf{35}(1),  857--876 (2021)

\bibitem{luo2022siman}
Luo, C., Jin, L., Chen, J.: Siman: exploring self-supervised representation learning of scene text via similarity-aware normalization. In: Proceedings of the IEEE/CVF CVPR. pp. 1039--1048 (2022)

\bibitem{marti2002iam}
Marti, U.V., Bunke, H.: The iam-database: an english sentence database for offline handwriting recognition. Int J Doc Anal Recogn  \textbf{5},  39--46 (2002)

\bibitem{evaluating_michael_2019}
Michael, J., Labahn, R., Grüning, T., Zöllner, J.: Evaluating sequence-to-sequence models for handwritten text recognition. IEEE ICDAR  (2019)

\bibitem{muehlberger2019transforming}
Muehlberger, G., Seaward, L., Terras, M., Oliveira, S.A., Bosch, V., Bryan, M., Colutto, S., D{\'e}jean, H., Diem, M., Fiel, S., et~al.: Transforming scholarship in the archives through handwritten text recognition: Transkribus as a case study. Journal of documentation  \textbf{75}(5),  954--976 (2019)

\bibitem{nikitha2020handwritten}
Nikitha, A., Geetha, J., JayaLakshmi, D.: Handwritten text recognition using deep learning. In: 2020 RTEICT. pp. 388--392. IEEE (2020)

\bibitem{noroozi2016unsupervised}
Noroozi, M., Favaro, P.: Unsupervised learning of visual representations by solving jigsaw puzzles. In: ECCV. pp. 69--84. Springer (2016)

\bibitem{novotny2018self}
Novotny, D., Albanie, S., Larlus, D., Vedaldi, A.: Self-supervised learning of geometrically stable features through probabilistic introspection. In: Proceedings of the IEEE CVPR. pp. 3637--3645 (2018)

\bibitem{ozbulak2023know}
Ozbulak, U., Lee, H.J., Boga, B., Anzaku, E.T., Park, H., Van~Messem, A., De~Neve, W., Vankerschaver, J.: Know your self-supervised learning: A survey on image-based generative and discriminative training. arXiv preprint arXiv:2305.13689  (2023)

\bibitem{souibgui2023text}
Souibgui, M.A., Biswas, S., Mafla, A., Biten, A.F., Forn{\'e}s, A., Kessentini, Y., Llad{\'o}s, J., Gomez, L., Karatzas, D.: Text-diae: a self-supervised degradation invariant autoencoder for text recognition and document enhancement. In: proceedings of the AAAI conference on artificial intelligence. vol.~37, pp. 2330--2338 (2023)

\bibitem{offline_tay_2001}
Tay, Y.H., Lallican, P.M., Khalid, M., Viard-Gaudin, C., Knerr, S.: Offline handwritten word recognition using a hybrid neural network and hidden markov model  (2001)

\bibitem{TransformerVaswani}
Vaswani, A., Shazeer, N., Parmar, N., Uszkoreit, J., Jones, L., Gomez, A.N., Kaiser, L.u., Polosukhin, I.: Attention is all you need. In: Guyon, I., Luxburg, U.V., Bengio, S., Wallach, H., Fergus, R., Vishwanathan, S., Garnett, R. (eds.) Advances in Neural Information Processing Systems. vol.~30. Curran Associates, Inc. (2017)

\bibitem{VidaToselliRiosVilaCalvoZaragoza:PR:2023}
Vidal, E., Toselli, A.H., Ríos-Vila, A., Calvo-Zaragoza, J.: End-to-end page-level assessment of handwritten text recognition. Pattern Recognition  \textbf{142},  109695 (2023)

\bibitem{yamaguchi2021image}
Yamaguchi, S., Kanai, S., Shioda, T., Takeda, S.: Image enhanced rotation prediction for self-supervised learning. In: 2021 IEEE International Conference on Image Processing. pp. 489--493. IEEE (2021)

\bibitem{yang2022reading}
Yang, M., Liao, M., Lu, P., Wang, J., Zhu, S., Luo, H., Tian, Q., Bai, X.: Reading and writing: Discriminative and generative modeling for self-supervised text recognition. In: Proceedings of the 30th ACM International Conference on Multimedia. pp. 4214--4223 (2022)

\bibitem{yang2022fully}
Yang, Z., Yu, H., He, Y., Sun, W., Mao, Z.H., Mian, A.: Fully convolutional network-based self-supervised learning for semantic segmentation. IEEE Trans Neural Networks Learn Syst  (2022)

\bibitem{zhang2022cmt}
Zhang, X., Wang, J., Jin, L., Ren, Y., Xue, Y.: Cmt-co: Contrastive learning with character movement task for handwritten text recognition. In: Proceedings of the ACCV. pp. 3104--3120 (2022)

\bibitem{zhang2022chaco}
Zhang, X., Wang, T., Wang, J., Jin, L., Luo, C., Xue, Y.: Chaco: Character contrastive learning for handwritten text recognition. In: ICFHR. pp. 345--359. Springer (2022)

\end{thebibliography}






\clearpage

\begin{appendix}


\section{Experimental set-up}

In this section we expand the experimental configuration considered in the work. More precisely, we detail the data augmentation processes, the visual encoder architecture, and the partitioning of the RIMES dataset used.

\subsection{Data augmentation}
We leverage hard data augmentation techniques enabled by the SSL methods employed, complemented by additional augmentation strategies inspired by seminal works in the field \cite{aberdam2021sequence, zhang2022chaco}. This approach results in the adoption of the following transformations by using the \emph{imgaug} augmentation package \cite{jung2020imgaug}:

\begin{lstlisting}[language=Python, caption=Data augmentation used in the SSL stage.]
aug = iaa.Sequential([
    iaa.SomeOf((1, 8), [
        # Geometric
        iaa.ScaleX((0.75, 1.0), mode='constant', cval=255),
        iaa.ScaleY((0.75, 1.25), mode='constant', cval=255),
        iaa.TranslateX(percent=(-0.1, 0.1), mode='constant', cval=255),
        iaa.TranslateY(percent=(-0.15, 0.15), mode='constant', cval=255),
        iaa.Rotate((-5., 5.), mode='constant', cval=255),
        iaa.ShearX((-25, 25), mode='constant', cval=255),
        iaa.PiecewiseAffine(scale=(0.02, 0.04)),
        # Color
        iaa.Sharpen(alpha=(0.1, 0.5), lightness=(0.75, 1.25)),
        iaa.LinearContrast((0.5, 1.5)),
        # Blur
        iaa.MotionBlur(k=(5, 7)),
        iaa.GaussianBlur((1.0, 2.0)),
        # Noise
        iaa.AdditiveGaussianNoise(scale=(1, 25))
        ], random_order=False)
])
# Morphological
num = random.uniform(0, 1)
if num < 0.33:
    image = erode(image)
elif random.uniform(0, 1) < 0.66:
    image = dilate(image)
image = aug(image=image)
\end{lstlisting}

\begin{figure}[!h]
  \centering
    \includegraphics[width=.6\linewidth]{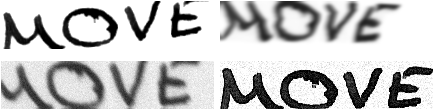}
    \caption{Examples of data augmentations.}
    \label{fig:data_aug}
\end{figure}

Examples of the transformations generated by the previous code are shown in Fig. \ref{fig:data_aug}.

Furthermore, drawing on established practices in the literature, we also employ the data augmentation strategy during the supervised stage. This includes the previous data augmentations excluding hard geometric augmentations.

\subsection{Visual encoder architectural details}

In this subsection, Table \ref{tab:cnn_details} details the precise configuration of the encoder architecture contemplated in the work. Note that, as commented in the manuscript, this scheme stands for a ResNet-29 architecture \cite{cheng2017focusing} with slight modifications.

\begin{table}[!ht]
\centering
\caption{Architectural details of the neural encoder considered. Block refers to a block of convolutional layers.}
\label{tab:cnn_details}
\begin{tabular}{lcccc}
\toprule[1pt]
\multirow{2}{*}{\textbf{Layer}} & \multicolumn{4}{c}{\textbf{Configuraton}} \\ \cline{2-5} 
 & \multicolumn{1}{c}{Kernel, Channels} & \multicolumn{1}{l}{} & Stride & Padding \\ \hline
Convolution & $3\times3$, 32 &  & (1, 1) & same \\
Convolution & $3\times3$, 64 &  & (1, 1) & same \\
Max pooling & $2\times2$, - &  & (2, 2) & (0, 0) \\
Block & $\begin{bmatrix}
3\times3, & 128\\
3\times3, & 128
\end{bmatrix}$ & $\times1$ & (1, 1) & same \\
Max pooling & $2\times2$, - &  & (2, 2) & (0, 0) \\
Block & $\begin{bmatrix}
3\times3, & 256\\
3\times3, & 256
\end{bmatrix}$ & $\times2$& (1, 1) & same \\
Max pooling & $2\times2$, - &  & (2, 2) & (0, 0) \\
Block & $\begin{bmatrix}
3\times3, & 512\\
3\times3, & 512
\end{bmatrix}$ & $\times5$ & (1, 1) & same \\
Convolution & $3\times3$, 512 &  & (1, 1) & same \\
Max pooling & $2\times2$, - &  & (2, 2) & (0, 0) \\
Block & $\begin{bmatrix}
3\times3, & 512\\
3\times3, & 512
\end{bmatrix}$ & $\times3$ & (1, 1) & same \\
Max pooling & $2\times2$, - &  & (2, 2) & (0, 0) \\
Convolution & $3\times3$, 512 &  & (1, 1) & same \\
\bottomrule[1pt]
\end{tabular}
\end{table}

\subsection{RIMES partition details}

In this work, we utilized the latest version of the RIMES word dataset \cite{grosicki2024rimes}. This dataset is organized into 57 folders, each referred to as a block, containing each one 100 letters represented as word-level images. To our best knowledge, there is no official partitioning scheme for this version of the dataset, nor are there prior studies that use it. Consequently, we have created arbitrary partitions for this work whose details are now provided to ensure reproducibility and facilitate future comparisons:

\begin{table}[!h]
\centering
\caption{Table showing the partitions employed in RIMES dataset}
\begin{tabular}{llcc}
\toprule[1pt]
\textbf{Partition} &  & \textbf{Blocks} & \textbf{Samples} \\ \cline{1-1} \cline{3-4} 
Train &  & 1-34 & 139490 \\
Validation &  & 35-39 & 20217 \\
Test &  & 40-57 & 68651 \\
\bottomrule[1pt]
\end{tabular}
\end{table}

\end{appendix}

\end{document}